\documentclass[review]{elsarticle}

\usepackage{lineno,hyperref}
\modulolinenumbers[5]

\usepackage{amssymb}
\usepackage{amsmath}
\usepackage{psfrag}
\usepackage{epsfig}
\usepackage{graphics}
\usepackage{graphicx}
\usepackage{color}
\usepackage{subfigure}
\usepackage{multirow}
\usepackage{threeparttable}
\usepackage{booktabs}
\usepackage{bbm}

\usepackage{algorithm}
\usepackage{algpseudocode}

\usepackage[T1]{fontenc}

\journal{Journal of \LaTeX\ Templates}
\biboptions{authoryear}

\begin{document}

\begin{frontmatter}



\title{Two-Stream Regression Network for Dental Implant Position Prediction}

\author[label1,label2,label3]{Xinquan Yang}
\ead{yangxinquan2021@email.szu.edu.cn}
\author[label4]{Xuguang Li}
\ead{lixuguang@szu.edu.cn}
\author[label1,label2,label3]{Xuechen Li}
\ead{timlee@szu.edu.cn}
\author[label5]{Wenting Chen}
\ead{wentichen7-c@my.cityu.edu.hk}
\author[label1,label2,label3]{Linlin~Shen\corref{mycorrespondingauthor}}
\ead{llshen@szu.edu.cn}
\author[label4]{Xin Li}
\ead{lizn007@126.com}
\author[label4]{Yongqiang Deng}
\ead{qiangyongdeng@sina.com}
\cortext[mycorrespondingauthor]{Corresponding author}

\address[label1]{College of Computer Science and Software Engineering, Shenzhen University, Shenzhen, China}
\address[label2]{AI Research Center for Medical Image Analysis and Diagnosis, Shenzhen University, Shenzhen, China}
\address[label3]{National Engineering Laboratory for Big Data System Computing Technology, Shenzhen University, China}
\address[label4]{Department of Stomatology, Shenzhen University General Hospital, Shenzhen, China}
\address[label5]{Department of Electrical Engineering, City University of Hong Kong, Hongkong, China}



\begin{abstract}
In implant prosthesis treatment, the design of the surgical guide heavily relies on the manual location of the implant position, which is subjective and prone to doctor's experiences. When deep learning based methods has started to be applied to address this problem, the space between teeth are various and some of them might present similar texture characteristic with the actual implant region. Both problems make a big challenge for the implant position prediction. In this paper, we develop a two-stream implant position regression framework (TSIPR), which consists of an implant region detector (IRD) and a multi-scale patch embedding regression network (MSPENet), to address this issue. For the training of IRD, we extend the original annotation to provide additional supervisory information, which contains much more rich characteristic and do not introduce extra labeling costs. A multi-scale patch embedding module is designed for the MSPENet to adaptively extract features from the images with various tooth spacing. The global-local feature interaction block is designed to build the encoder of MSPENet, which combines the transformer and convolution for enriched feature representation. During inference, the RoI mask extracted from the IRD is used to refine the prediction results of the MSPENet. Extensive experiments on a dental implant dataset through five-fold cross-validation demonstrated that the proposed TSIPR achieves superior performance than existing methods.
\end{abstract}



\begin{keyword}
Implant Prosthesis \sep Dental Implant \sep Vision Transformer \sep Deep Learning



\end{keyword}

\end{frontmatter}


\section{Introduction}\label{sec1}
Dental implant is a common surgical procedure in oral and maxillofacial surgery~\citep{varga2020guidance}, in which the surgical guide plays an important role in precise bone drilling and implant placement~\citep{gargallo2021intra,vinci2020accuracy}. However, the design of the surgical guide heavily relies on the manual location of the implant position using the patient's panoramic radiographic image, or cone beam computed tomography (CBCT) data, which is subjective and prone to doctor's experiences~\citep{liu2021transfer}. In contrast, artificial intelligence (AI) methods can quickly locate the implant position, which is trained using a large number of successful implant cases designed by dentists with rich related clinical experiences. As the AI methods always give the same prediction for the same data, it is thus more objective when predicting the implant position and have less varitions. Therefore, it inspire us to improve the efficiency of surgical guide design using deep learning-based methods.

Generally, the prediction of implant location in CBCT data can be considered as a three-dimensional (3D) regression task. However, the training of a 3D neural network requires a lot of training data, which leads to higher collection and labeling costs. The common solution is to convert the 3D CBCT data into a series of 2D slices. Dental-YOLO~\citep{widiasri2022dental} utilized the 2D sagittal view of CBCT to measure the oral bone, e.g., the alveolar bone, and determine the implant position indirectly. ImplantFormer~\citep{yang2022implantformer} predicts the implant position using the 2D axial view of tooth crown images and projects the prediction results back to the tooth root by the space transform algorithm. Even though current methods can achieve great performance on the implant position prediction, these methods do not consider the influence of the variation of the tooth cross-sectional area, which may degrade the performance of the prediction network.

\begin{figure}
\centering
\includegraphics[width=1.0\linewidth]{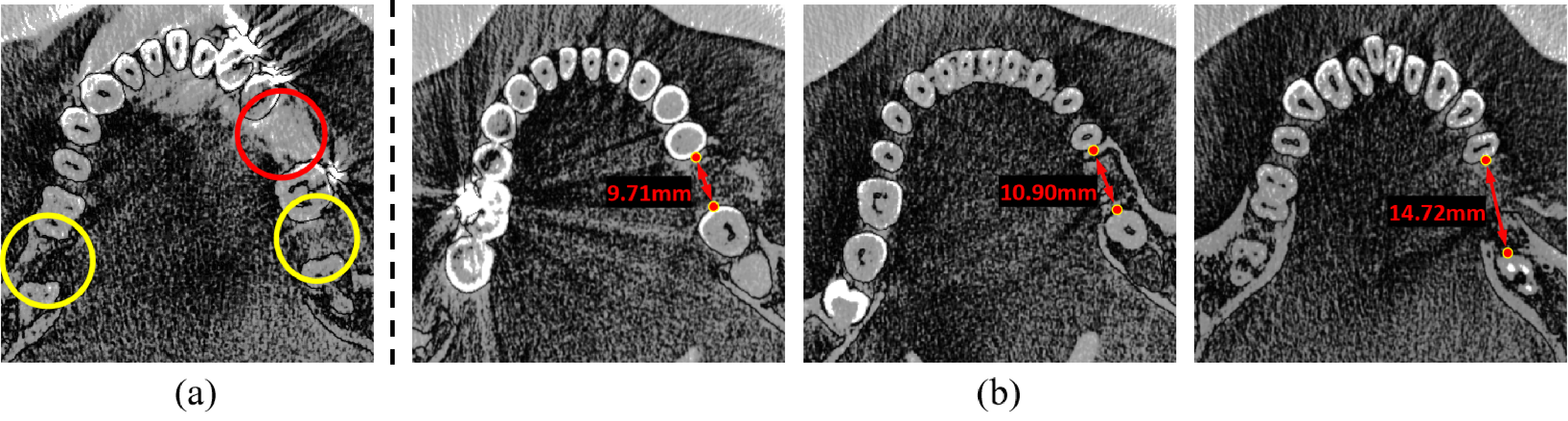}
\caption{(a) Example images of the sparse teeth in the tooth crown image. The red and yellow circles denote the actual regression region and the area prone to generate false alarms, respectively. (b) Comparison of the tooth crown images with different tooth spacing.} \label{fig_patch_sparse}
\end{figure}

First of all, physically, the irregular structure of the tooth leads to the decrease of cross-sectional area from the tooth crown to the tooth root. As a result, the gap between neighboring teeth increase as the number of CT layers grows. When the gap between neighboring teeth is big enough, the regions between sparse teeth may have a similar characteristic with the actual implant region (see Fig.~\ref{fig_patch_sparse}(a)), which will misguide the prediction network to generate false positive detection. Secondly, as shown in Fig.~\ref{fig_patch_sparse}(b), the tooth spacing has a big variation (from 9.71 to 14.72 mm) across different patients,  where the fixed kernel size of convolution or patch embedding can not extract robust features. Both problems make a big challenge for implant position regression.

To tackle these challenges, we develop a two-stream implant position regression framework (TSIPR), which consists of an implant region detector (IRD) and a multi-scale patch embedding regression network (MSPENet). IRD is an object detector designed to locate the implant region and filter out the region of sparse teeth. The training of IRD uses the extended bounding box of implant position annotation. Compared to the ground-truth position (the red point in Fig.~\ref{fig_extend}) that has little useful texture, the extended box (the dashed blue box in Fig.~\ref{fig_extend}) contains much more rich characteristics, i.e., the neighboring teeth. More importantly, the acquisition of the extended box do not introduce extra labeling costs. MSPENet is devised to regress the precise implant position. To adaptively extract features from the images with various tooth spacing, we design a multi-scale patch embedding module, to aggregate the features extracted from different sizes of patch embedding for more robust features. A global-local feature interaction block (GLFIB) is designed as the encoder of MSPENet, which integrates the global context of the transformer and the local texture extracted from the convolution for enriched feature representation. As the output of IRD is the bounding box with the high confidence, which represents the most probable implant region, the false detection generated at the other regions will be removed. During inference, the heatmap with implant position generated by the MSPENet is multiplied with the region of interest (RoI) mask extracted from the IRD, to refine the prediction results. The main contributions of this work are summarized as follows:
\begin{itemize}
\item We develop a two-stream implant position regression framework (TSIPR), which includes a MSPENet for implant position regression and an IRD to locate the most probable implant region. The RoI mask extracted from the IRD is used to refine the prediction results of the MSPENet, which greatly reduces the false detection rate.
\item For the training of IRD, we extend the original annotation to provide additional supervisory information. The extended bounding box contains much more rich characteristic and do not introduce extra labeling costs.
\item For the MSPENet, a multi-scale patch embedding module is designed to adaptively extract features from the images with various tooth spacing and a global-local feature interaction block is designed to integrate the global context and the local texture for enriched feature representation.
\item Extensive experiments on a dental implant dataset demonstrates that the proposed TSIPR achieves superior performance than the existing methods, especially for patients with sparse teeth.
\end{itemize}

The rest of the paper is organized as follows. Section 2 briefly reviews the related works. Section 3 gives the details of the proposed method. Section 4 presents experiments on a dental implant dataset and the experimental results are compared with that of mainstream detectors and the state-of-the-art methods. Section 5 provides the conclusions.

\section{Related work}\label{sec2}
\subsection{Deep learning in dentistry}\label{subsec2}
Deep learning technology has been applied in many tasks of dentistry, such as tooth segmentation, orthodontic treatment, and dental implant classification. For tooth segmentation, the studies mainly focus on two kinds of data, i.e. CBCT data and 3D dental point cloud data. Jang et al.~\citep{jang2021fully} proposed a fully automated method of identifying and segmenting 3D individual teeth from dental CBCT images, which addressed the difficulty of separating the individual tooth from adjacent teeth and its surrounding alveolar bone. Zhang et al.~\citep{zhang2018effective} proposed a label tree-based method to assign each tooth several labels and decompose the segmentation task into several sub-tasks for resolving the problem of limited training data. Mahdi et al.~\citep{mahdi2020optimization} proposed a residual network-based faster R-CNN model for automatic teeth recognition, which further refined the candidates using a candidate optimization technique that evaluates both positional relationship and confidence score. For orthodontic treatment, Qian et al.~\citep{qian2020cephann} proposed a multi-head attention neural network for detecting cephalometric landmarks, which consists of a multi-head and an attention. The multi-head component adopts multi-head subnets to learn different features from various aspects. The attention uses a multi-attention mechanism to refine the detection based on the features extracted by the multi-head. Dai et al.~\citep{dai2019locating} proposed a new automated cephalometric landmark localization method based on GAN, which trains an adversarial network to learn the mapping from features to the distance map of a specific target landmark. For the task of dental implants classification, Sukegawa et al.~\citep{sukegawa2020deep} evaluated a series of CNN models, i.e. a basic CNN with three convolutional layers, VGG16 and VGG19 transfer-learning models, and fine-tuned VGG16 and VGG19 for implant classification. Kim et al.~\citep{kim2020transfer} developed an optimal pre-trained network architecture for identifying four different types of implants, i.e. Brånemark Mk TiUnite Implant, Dentium Implantium Implant, Straumann Bone Level Implant and Straumann Tissue Level Implant on intraoral radiographs.

\begin{figure}
\centering
\includegraphics[width=0.7\linewidth]{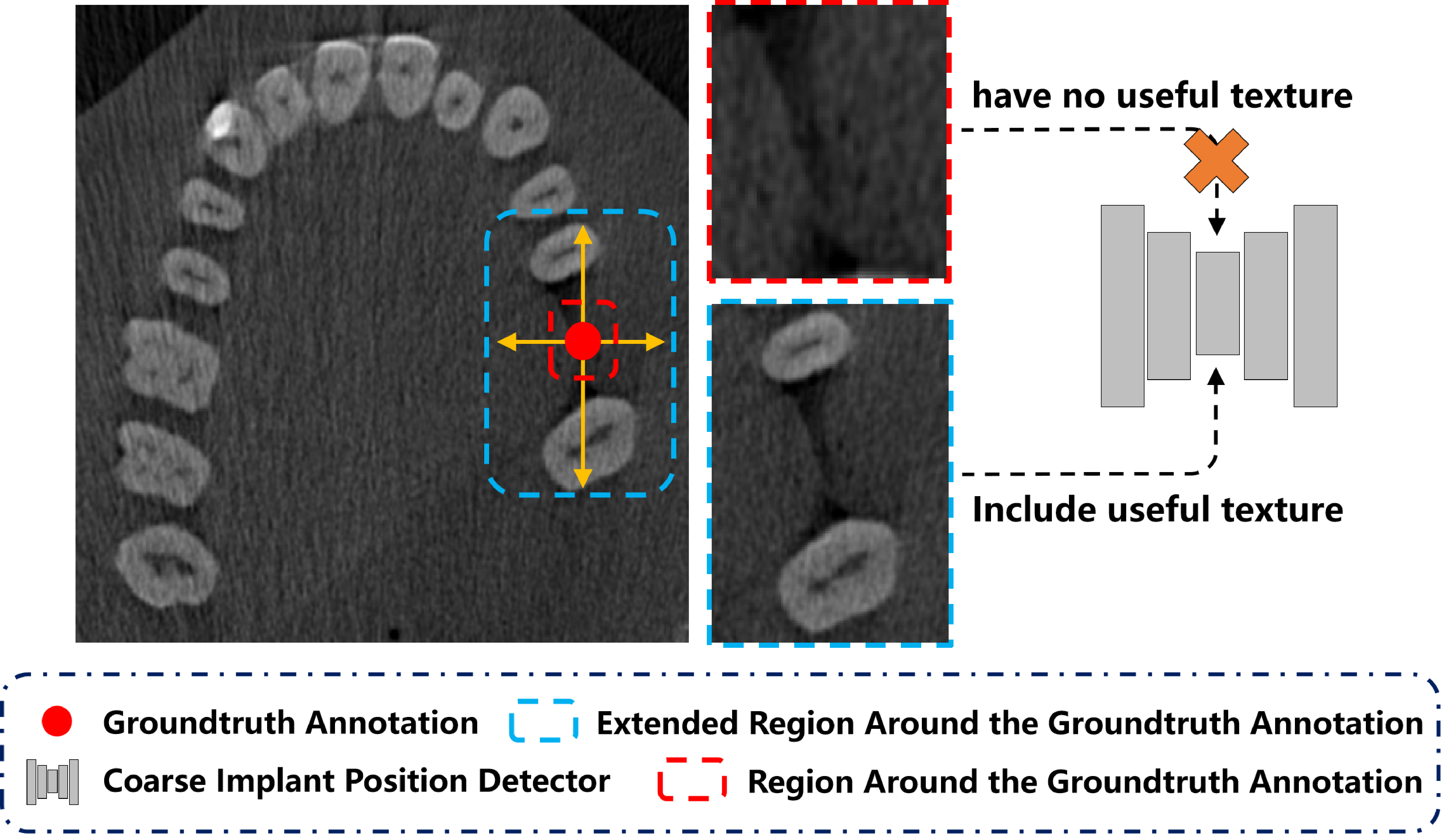}
\caption{Comparison of implant position annotation and the extended implant region.} \label{fig_extend}
\end{figure}
\subsection{Deep learning in object detection}\label{subsec2}
Current object detectors can be divided into two categories, i.e. anchor-based and anchor-free. The anchor-based detector sets the pre-defined anchor box before training and the anchor-free detector directly regresses the bounding box of the object. The anchor-based detector can be further grouped into one-stage and two-stage methods. SSD~\citep{liu2016ssd} and YOLO~\citep{redmon2016you} are classic one-stage detectors, which directly predict the bounding box and category of objects based on the feature maps. Faster R-CNN~\citep{ren2015faster} is a classical two-stage detector that consists of a region proposal network (RPN) and a prediction network (R-CNN~\citep{girshick2014rich}). A series of detection algorithms ~\citep{cai2018cascade,sun2021sparse,tan2020efficientdet,wang2022yolov7} have been proposed to improve the performance of these anchor-based detectors. Compared to the anchor-based detector that heavily relies on the predefined anchor box, the anchor-free detector breaks such limitation. CornerNet~\citep{law2018cornernet} simplified the prediction of the object bounding box as the regression of the top-left corner and the bottom-right corner. CenterNet~\citep{duan2019centernet} further simplified CornerNet by regressing the center of object. With the development of the vision transformer, transformer-based anchor-free detector achieves great success in object detection. DETR~\citep{carion2020end} employs ResNet as the backbone and introduces a transformer-based encoder-decoder architecture for the object detection task. Deformable DETR~\citep{zhu2020deformable} extends DETR with sparse deformable attention that reduces the training time significantly.

\subsection{Deep learning in implant position estimation}\label{subsec3}
The computer-aided diagnosis (CAD) systems has been applied to dental implant planning. Sadighpour et al.~\citep{sadighpour2014application} developed an ANN model which utilized a number of input factors to formulate a decision regarding the type of prosthesis (fixed or removable) and the specific design of the prosthesis for rehabilitation of the edentulous maxilla. Lee et al.~\citep{lee2012development} applied fuzzy recognition map for implant abutment selection. Szejka et al.~\citep{szejka2011reasoning} developed an interactive reasoning system which requires the dentist to select the region of interest within a 3D bone model based on computed tomography (CT) images, to help the selection of the optimum implant length and design. However, these CAD systems need manual hyperparameter adjustment.

Recently, researchers proposed different approaches to determine the implant position using the panoramic radiographic images and 2D slices of CBCT. Kurt et al.~\citep{kurt2021deep} utilised multiple pre-trained convolutional networks to segment the teeth and jaws to locate the missing tooth and generate a virtual tooth mask according to the neighbouring teeth' location and tilt. Widiasri et al. introduced Dental-YOLO~\citep{widiasri2022dental} to detect the alveolar bone and mandibular canal based on the sagittal view of CBCT to determine the height and width of the alveolar bone. Yang et al.~\citep{yang2022implantformer} developed a transformer-based implant position regression network (ImplantFormer), which directly predicts the implant position on the 2D axial view of tooth crown images and projects the prediction results back to the tooth root by the space transform algorithm. However, these methods do not consider the irregular structure of tooth, which is a big challenge to produce false alarms.

\begin{algorithm}
\caption{Pseudocode of the workflow of TSIPR.}\label{algorithm}
\begin{algorithmic}[1]
\Require
    The tooth crown image of the patient $\tilde{p}$ - $I_{\tilde{p}}$.
\Ensure
    The implant position at tooth root $y_r$.
\State $M=IRD(I_{\tilde{p}})$
\State $H=MSPENet(I_{\tilde{p}})$
\State $\hat H=M\otimes H$
\State $Pos^{\tilde{p}}_c(i)=Extract(\hat H)$
\State $Pos^{\tilde{p}}_r(j)=T_{Pos^{\tilde{p}}_c \to Pos^{\tilde{p}}_r}(Pos^{\tilde{p}}_c(i))$
\end{algorithmic}
\end{algorithm}

\section{Method}\label{sec4}
\begin{figure}
\centering
\includegraphics[width=1.0\linewidth]{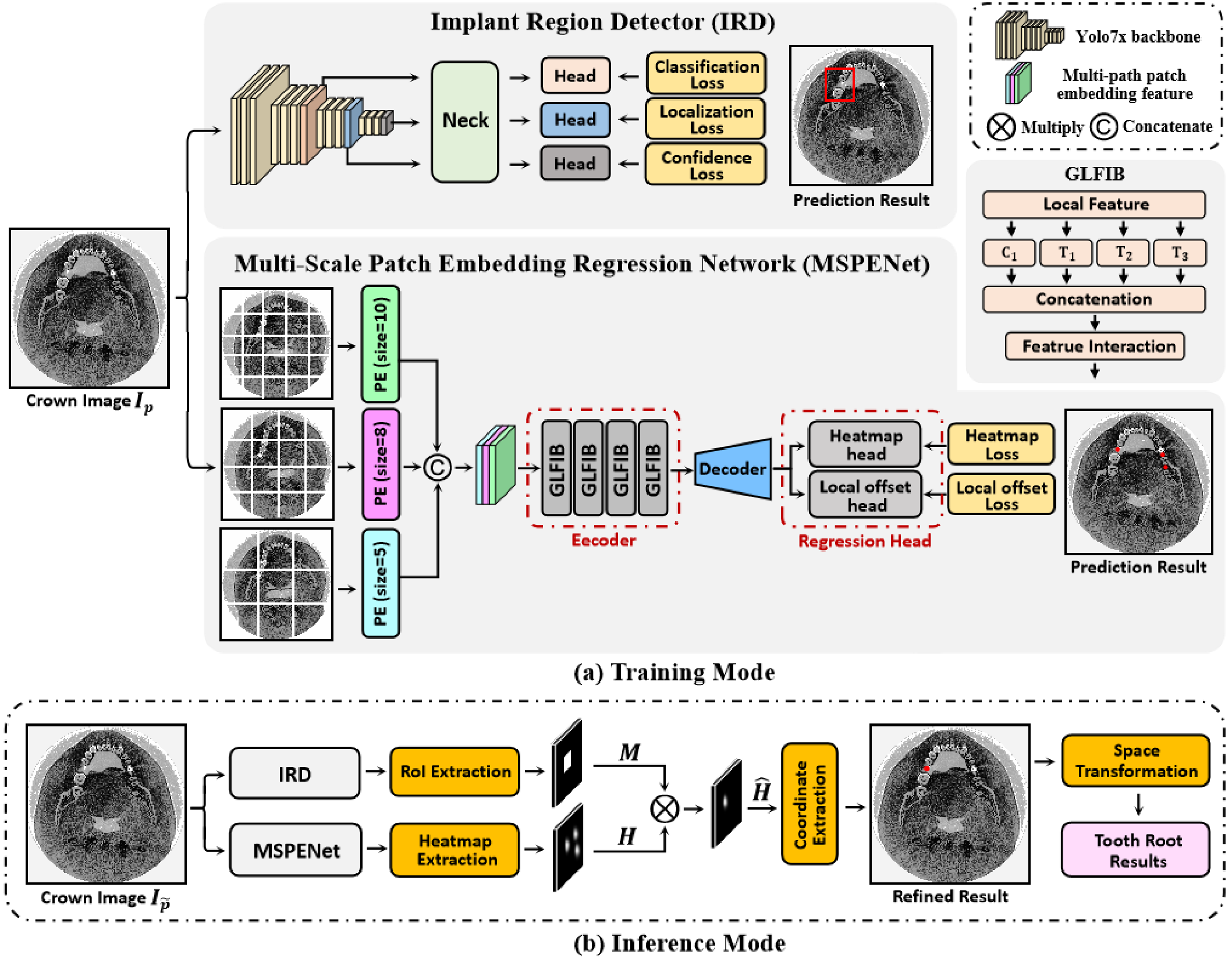}
\caption{The overview of the proposed TSIPR.} \label{fig_network}
\end{figure}

Using tooth crown image to regress the implant position has been shown to be effective in~\citep{yang2022implantformer}. Therefore, in this work, we follow this paradigm to train TSIPR. An overview of TSIPR is presented in Fig.~\ref{fig_network}. It mainly consists of an implant region detector (IRD) and a multi-scale patch embedding regression network (MSPENet). We provide a pseudocode as Algorithm~\ref{algorithm} to explain the workflow of TSIPR. During training, IRD and MSPENet are trained separately. In inference, IRD and MSPENet share the same tooth crown image $I_{\tilde{p}}\in \mathbb{R}^{H\times W\times C}$ of patient $\tilde{p}$ as input, and the outputs of IRD and MSPENet are the most probable implant region $M\in \mathbb{R}^{\frac{H}{4}\times \frac{W}{4}}$ and the heatmap with the implant position $H\in \mathbb{R}^{\frac{H}{4}\times \frac{W}{4}}$, respectively. Then, we multiply $M$ with $H$ to filter out the error detection generated by the MSPENet and obtain the filtered heatmap $\hat H\in \mathbb{R}^{\frac{H}{4}\times \frac{W}{4}}$. In the end, the refined implant positions $Pos^{\tilde{p}}_c(i)=(x^{\tilde{p}}_i,y^{\tilde{p}}_i,z^{\tilde{p}}_i)$ extracted from $\hat H$ are projected from the tooth crown to tooth root area by the space transformation algorithm $T_{Pos^{\tilde{p}}_c \to Pos^{\tilde{p}}_r}$~\citep{yang2022implantformer} to obtain the implant position at tooth root$Pos^{\tilde{p}}_r(j)=(\hat x^{\tilde{p}}_j,\hat y^{\tilde{p}}_j,\hat z^{\tilde{p}}_j)$.





\subsection{Implant Region Detector}
As shown in Fig.~\ref{fig_patch_sparse}(a), the gap between neighboring teeth at the region of sparse teeth has a similar characteristic with the actual implant region, which will misguide the prediction network to generate false detection. To tackle this problem, we propose to train an implant region detector (IRD) to filter out the false detection.

The IRD is trained using the extended bounding box of implant position annotation (shown as the dashed blue box in Fig.~\ref{fig_extend}). Different from the original implant position annotation (the red point in Fig.~\ref{fig_extend}) that has little useful texture at the implant region, the extended box includes the neighboring teeth that enable the implant region to contain rich characteristics. Moreover, at the scale of the extended region, the real implant region has a larger interval between the neighboring teeth than that between sparse teeth. Both characteristics can be easily captured by the IRD. As the output of IRD is the bounding box with the highest confidence, which represents the most probable implant region, the false detection generated at the region of sparse teeth will be removed. Specifically, we set the size of the extended box as $128\times128$ to ensure that the texture of neighboring teeth is included. The extended bounding box will not introduce additional labeling costs, as the coordinate of the extended box is determined according to the original annotation.

Considering that the output of IRD is used to refine the detection results, a trade-off between location performance and inference speed is required. Therefore, we introduce a strong detector, i.e. YOLOv7-X~\citep{wang2022yolov7} as IRD. The network architecture is shown in Fig.~\ref{fig_network}, which consists of a backbone, a neck and three prediction heads. IRD takes $I_p$ as input. Feature maps of three different resolutions are extracted by the backbone and then input into the neck for feature fusion. Finally, the location prediction head generates the bounding box of the probable implant region. The IRD network is optimized by three loss functions, i.e. classification loss $\mathcal{L}_{cls}$, localization loss $\mathcal{L}_{loc}$ and confidence loss $\mathcal{L}_{cof}$. $\mathcal{L}_{cls}$ and $\mathcal{L}_{cof}$ are the cross-entropy loss function and $\mathcal{L}_{loc}$ is the GIoU loss~\citep{rezatofighi2019generalized}. The overall training loss of IRD is:
\begin{equation}
\mathcal{L}_I=\mathcal{L}_{cls}+\mathcal{L}_{loc}+\mathcal{L}_{cof}
\end{equation} \label{eq1}

\subsection{Multi-scale Patch Embedding Regression Network}
In implant position regression, ViT relies on the patch embedding operation and multi-head self-attention (MHSA) to build the relationship between the implant position and the texture of neighboring teeth~\citep{yang2022implantformer}. However, due to the structural difference in the patient's mouth, the teeth spacing of tooth crown image has a big variation in different patients (see Fig.~\ref{fig_patch_sparse}(a)). The single kernel size of patch embedding in ViT can not perform well in this situation. Additionally, although ViT shows great performance in capturing global context~\citep{tuli2021convolutional}, it may ignore the local texture within each patch~\citep{lowe1999object}, which can not extract enriched feature representation. In contrast, convolutional neural network (CNN) benefits from the inductive bias to capture local texture~\citep{baker2018deep}. To tackle the above issues, we design a multi-scale patch embedding regression network (MSPENet) to predict the implant position, which mainly consists of three parts: i) Multi-scale Patch Embedding, ii) Encoder and Decoder, iii) Regression Head. An overview of the proposed network is presented in Fig.~\ref{fig_network}.

Given a tooth crown image $I_p$, the multi-scale patch embedding module firstly extracts robust features by three different sizes of patch embedding. Then, the output features are integrated together by concatenation and input into the encoder for further feature extraction. The decoder is used to recover the high-resolution representation from the output of encoder. In the end, the regression head aims to output a Gaussian heatmap that highlights a precise implant position. Next, we will introduce these modules in detail.

\subsubsection{Multi-scale Patch Embedding Module}
The multi-scale patch embedding module is devised to extract robust features from the input image. Similar to the CvT~\citep{wu2021cvt}, we use convolution with overlapping patches to implement the patch embedding layer. The convolutional patch embedding layer enables us to output features of the same resolution with different patch sizes. Specifically, we select three patch embedding layers with size of 5$\times$5, 8$\times$8, and 10$\times$10, respectively. The patch sizes are determined by the experimental results. The multi-scale patch embedding module takes $I_p^r$ as input and separately extracts image features in parallel. Then, the extracted multi-scale features are aggregated by concatenation. To fuse the features of different patch embedding layers, we use 1$\times$1 convolution to smooth the aggregated feature. The output feature is fed into the encoder for further feature learning.

\subsubsection{Encoder and Decoder}
Recent works show the benefit of combining convolution and transformer in network design~\citep{chen2022mobile,mehta2021mobilevit}, in which the convolution captures the local texture and the transformer extracts global context. Considering that the local texture within the patch is also important for the prediction of implant position, we devise a global-local feature interaction block (GLFIB) for Encoder to integrate both local texture and global context. The architecture of GLFIB is given in Fig.~\ref{fig_network}. GLFIB consists of three branches of transformer and one branch of convolution in parallel. We use multiple transformer modules to enrich the channel. This design aims to enable the network to focus more on capturing the relationship between different patches. To alleviate the computational burden, we follow~\citep{lee2022mpvit} to adopt depth-wise convolutions and the efficient factorized self-attention~\citep{xu2021co} to construct GLFIB. Specifically, the local feature of network $l\in \mathbb{R}^{\tilde{h}\times \tilde{w}\times \tilde{c}}$ is separately fed into each branch for feature extraction, and then the output features of branches are aggregated together by concatenation:
\begin{equation}
A=\textbf{concat}[C_1(l), T_1(l), T_2(l), T_3(l)],
\end{equation}
where $A\in \mathbb{R}^{\frac{\tilde{h}}{2}\times \frac{\tilde{w}}{2}\times 4\tilde{c}}$ is the aggregated feature. $C(\cdot)$ and $T(\cdot)$ is the convolution and transformer module, respectively. The kernel size of both modules are 3 ×3. After obtaining the aggregated features, we use $f(\cdot)$ to interact features between local texture and global context:
\begin{equation}
O=f(A),
\end{equation}
where $O\in \mathbb{R}^{\frac{\tilde{h}}{2}\times \frac{\tilde{w}}{2}\times 2\tilde{c}}$ is the final output feature. We use 1$\times$1 convolution with channel of $2\tilde{c}$ for $f(\cdot)$. The encoder of MSPENet consists of four cascaded GLFIB, and the output of the last GLFIB is used as input for the decoder.

The output of the Encoder is a high-level feature. To ensure fine-grained heatmap regression, three deconvolution layers are adopted as the Decoder to recover high-resolution features from the output of encoder. The Decoder consecutively upsamples feature map as high-resolution feature representations, in which the output resolution the same as the first GLFIB. In the end, the upsampled feature map is input into the regression network to locate the implant position.

\subsubsection{Regression Head}
The regression network consists of a heatmap head and a local offset head, which is used for predicting the implant position. The heatmap head generates an Gaussian heatmap $F\in[0,1]^{\frac{W}{g} \times \frac{H}{g}}$, where $g$ is the down sampling factor of the prediction and set as 4. Following the standard practice of CenterNet~\citep{zhou2019objects}, the ground-truth position is transformed into the heatmap $F$ using a 2D Gaussian kernel:
\begin{equation}
F_{xy}=\exp(-\frac{(x-{\tilde{t}_x})^2+(y-{\tilde{t}_y})^2}{2\sigma^2})
\end{equation}
where $(\tilde{t}_x,\tilde{t}_y)$ is ground-truth annotation in $F$. $\sigma$ is an object size-adaptive standard deviation  . The heatmap head is optimized by the focal loss~\citep{lin2017focal}:
\begin{equation}
\mathcal{L}_h=\frac{-1}{N}\sum_{xy}\left\{
\begin{array}{ccl}
(1-\hat F_{xy})^\alpha\log(\hat F_{xy}) & \text{if $F_{xy}=1$} \\
(1-\hat F_{xy})^\beta\log(\hat F_{xy})^\lambda\log(1-\hat F_{xy}) & \text{otherwise}
\end{array}\right.
\end{equation}
where $\alpha$ and $\beta$ are the hyper-parameters of the focal loss, $\hat F$ is the predicted heatmap. The heatmap $F$ is expected to be equal to 1 at the groundtruth position, and equal to 0 otherwise.

The local offset head computes the discretization error caused by $g$, which is used to further refine the prediction location. The loss of the local offset $L_o$ is optimized by L1 loss~\citep{girshick2014rich}. The overall training loss of MSPENet is:
\begin{equation}
\mathcal{L}_M=\mathcal{L}_h+\mathcal{L}_o
\end{equation}

\subsection{Two-Stream Implant Position Regression Framework}
TSIPR is proposed for better predicting the implant position, which parallelizes a coarse implant region detector - IRD and an implant position regression network - MSPENet. The output of IRD, i.e., a probable implant region, is used to filter out the error detection generated by the MSPENet. Specifically, in inference, given a tooth crown image $I_{\tilde{p}}$ of the patient $\tilde{p}$, the implant position can be derived as following procedure:

\begin{equation}
M=\textbf{IRD}(I_{\tilde{p}}^l,W_l),
\end{equation}
\begin{equation}
H=\textbf{MSPENet}(I_{\tilde{p}}^r,W_r),
\end{equation}
where $M\in \mathbb{R}^{\frac{\hat h}{4}\times \frac{\hat w}{4}}$ is the RoI mask of implant region. $W_l$ and $W_r$ represent the learning parameters of IRD and MSPENet, respectively. After getting the regression heatmap $H\in \mathbb{R}^{\frac{\hat h}{4}\times \frac{\hat w}{4}}$ from the MSPENet, to filter out the false positive predictions, the RoI mask is applied for refinement:
 \begin{equation}
y=E(M\otimes H),
\end{equation}
where $y$ is the extracted implant position at tooth crown and $E$ represents the coordinate extraction operation. $\otimes$ is the matrix multiplication operation. However, the predicted implant positions at the tooth crown area are not the real location of implant. To obtain the implant position at tooth root, we introduce a space transformation algorithm~\citep{yang2022implantformer}, which fit the center line of implant using predicted implant position at tooth crown and then extend the center line to the root area. By this means, the intersections of implant center line with 2D slices of tooth root image, i.e. the implant position at tooth root area, can be obtained.

\begin{figure}
\centering
\includegraphics[width=1.0\linewidth]{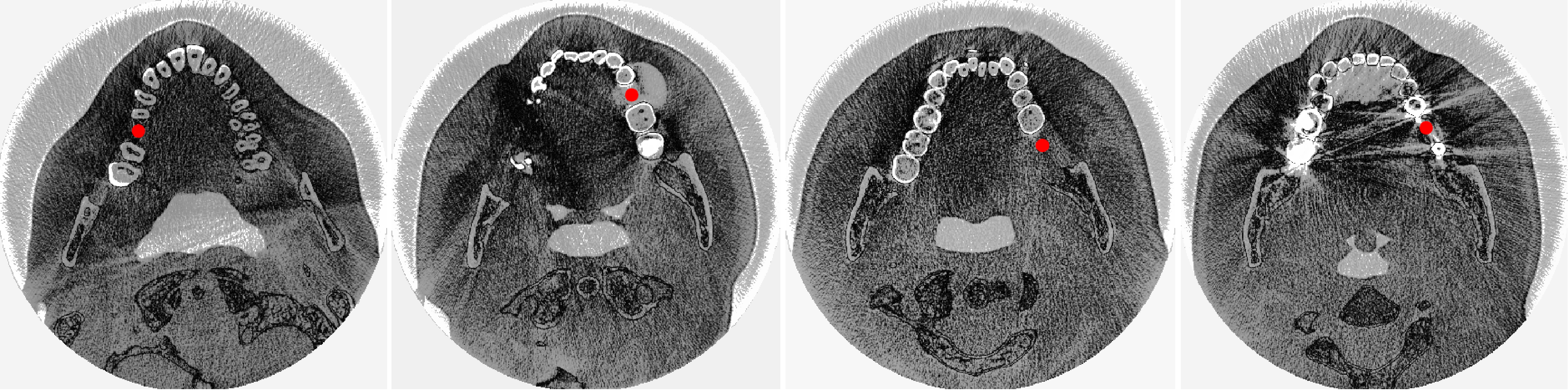}
\caption{Some sample images in the dental implant dataset. The red points denote the implant position annotation.} \label{fig_dataset}
\end{figure}

\section{Experiment}\label{sec6}
\subsection{Dataset}
The dental implant dataset used for evaluation is collected from the Shenzhen University General Hospital (SZUH), and all the implant positions were annotated by three experienced dentists. Specifically, the dataset contains 154 patients, from which 3045 2D slices of tooth crown are selected. Some sample images of the dataset are shown in Fig. 4. All the CBCT data were captured using the KaVo 3D eXami machine, supplied by Imagine Sciences International LLC. The resolution of the tooth crown image is $776\times776$. We perform five-fold cross-validation for the experiments, i.e. the number of the training and testing images in each fold is 2436 and 609, respectively.

Both the location and regression branches are trained using the same training image, while the training label and image size are different. The location branch - IRD uses the extended bounding box as annotation, which is generated by extending the ground-truth position to a fixed size ($128\times128$) box. The regression branch - MSPENet is trained using the original ground-truth position. The image size used in the location and regression branch are $640\times640$ and $512\times512$, respectively.

\subsection{Implementation Details}
Pytorch is used for model training and testing. For the training of IRD, we use a batch size of 16, SGD optimizer and a learning rate of 0.01. Four data augmentation methods, i.e. mosaic, mixup, photometric distortion and geometric are employed. The network is trained for 80 epochs. For the training of MSPENet, we use a batch size of 8, Adam optimizer and a learning rate of 0.0005 for network training. A series of augmentation methods, i.e. adding random noise~\citep{kaur2021automated}, enhancing contrast~\citep{ubhi2022neural}, random crop, random scale and random flip are employed. The network is trained for 140 epochs and the learning rate is divided by 10 at 40th and 60th epochs, respectively. All the models are trained and tested on the platform of TESLA V100 GPU. For the training of other baseline detectors, MMDetection library and ImageNet pre-training models are used.

\subsection{Evaluation Criteria}
In the clinical, the diameter of the implant is 3.5$\sim$5mm, and the mean error between the predicted and ideal implant position is required to be less than about 5 pixels (1mm with the CBCT imaging resolution in the paper), i.e., around 25\% of the size of implant. Therefore, instead of general $AP_{50}$, $AP_{75}$~\citep{kaur2022prediction} is used as the evaluation criteria in this work. The calculation of AP is defined as follows:
\begin{equation}
Precition=\frac{TP}{TP+FP}
\end{equation}
\begin{equation}
Recall=\frac{TP}{TP+FN}
\end{equation}
\begin{equation}
AP=\int^1_0P(r)dr
\end{equation}
Here TP, FP and FN are the number of correct, false and missed predictions, respectively. P(r) is the PR Curve where the recall and precision act as abscissa and ordinate, respectively.

\begin{table}
\centering
\caption{Performance Comparison of Different IRD.}\label{table_IRD}
\begin{tabular}{|c|c|c|c|l|}
\hline
Network   & Precision(\%) & Recall(\%) & $AP_{75}\%$  & FPS \\ \hline
Yolov7    & 78.1          & \textbf{84.5}       & 84.0 & \textbf{72}  \\ \hline
Yolov7-X  & \textbf{86.9} & 81.7       & 87.8 & 65  \\ \hline
Yolov7-W6 & 86.8          & 84.2       & \textbf{89.3} & 57  \\ \hline
\end{tabular}
\end{table}

\subsection{Performance Analysis}
\subsubsection{Comparison of different IRD}
IRD is designed for locating the implant region to refine the predicted implant position. Therefore, a high accurate detector with quick inference speed detector is required. To have a good trade-off between accuracy and speed for the IRD, we compare three versions of Yolov7 detector, results are listed in Table~\ref{table_IRD}. Since the implant region located by the IRD is a coarse area, the $AP_{50}$ is used as the evaluation criteria to assess the location performance. From the table we can observe that the Yolov7-X achieved the highest precision of 86.9\% and a medium AP value 87.8\%. Although the Yolov7-W6 achieved the best overall performance, i.e. 89.3\% AP, the precision rate is close to the Yolov7-X. Compared to the recall rate, the precision of the bounding box is a more important index for the IRD since only the highest confidence box is selected for each image. In terms of the inference time, Yolov7-W6 is slower than Yolov7-X for nearly 10 fps. Yolov7 has the highest inference speed, but the locating performance is poor. Consequently, we chose the Yolov7-X as the IRD for the implant region location task.

\begin{figure}
\centering
\includegraphics[width=1.0\linewidth]{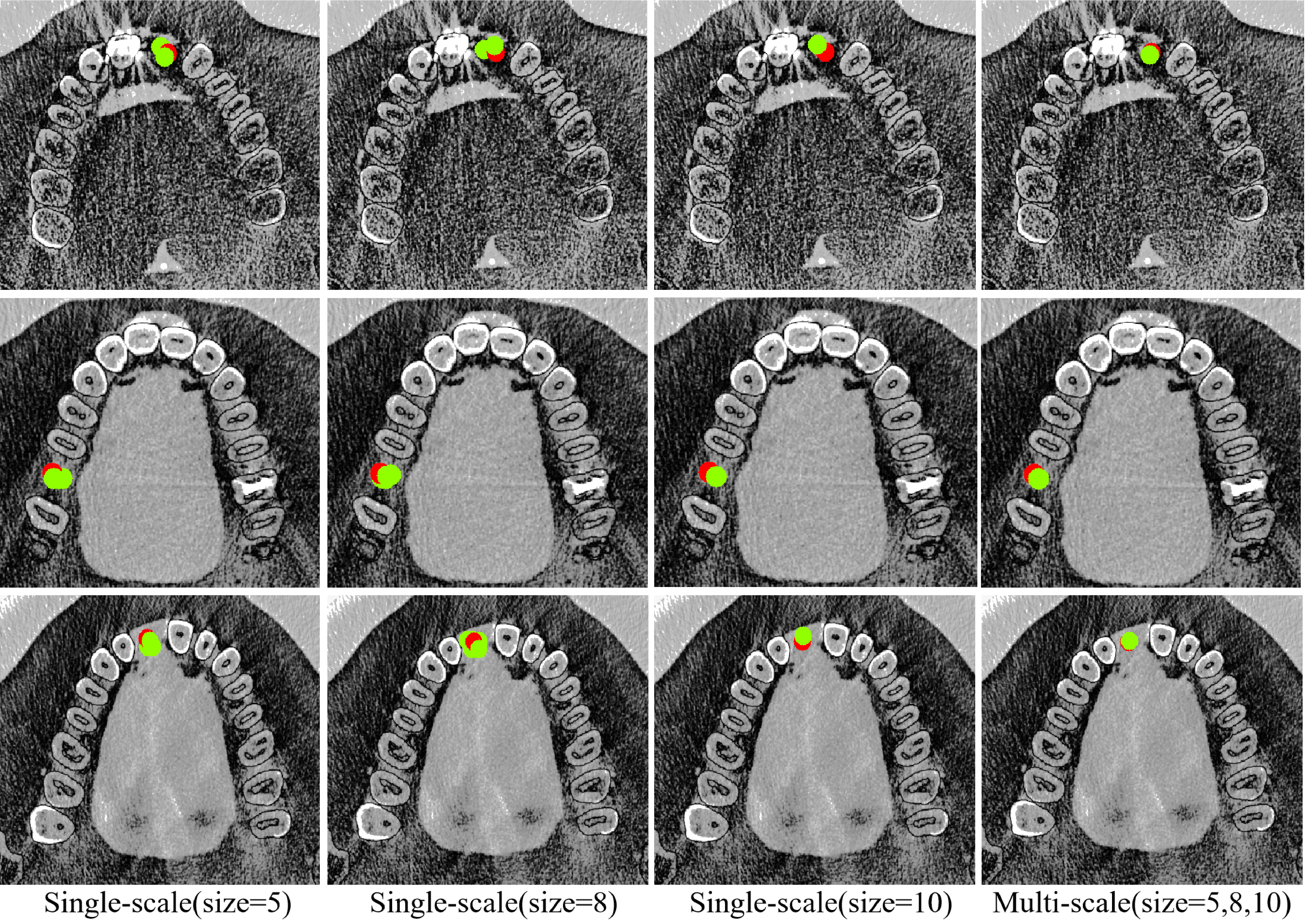}
\caption{Visual comparison of detection results between the single-scale and multi-scale patch embedding network. The red and green circles denote the ground-truth position and the predicted implant position, respectively.} \label{fig_exp_patch}
\end{figure}

\begin{table}[]
\centering
\caption{Ablation study of different patch size in the multi-scale patch embedding module.}\label{table_patchsize}
\begin{tabular}{|l|llllll|l|}
\hline
\multirow{2}{*}{Network} & \multicolumn{6}{l|}{Patch Size}                                                                                                 & \multirow{2}{*}{$AP_{75}\%$} \\ \cline{2-7}
                         & \multicolumn{1}{l|}{5} & \multicolumn{1}{l|}{6} & \multicolumn{1}{l|}{7} & \multicolumn{1}{l|}{8} & \multicolumn{1}{l|}{9} & 10 &                     \\ \hline
\multirow{7}{*}{MSPENet} & \multicolumn{1}{l|}{\checkmark}  & \multicolumn{1}{l|}{}  & \multicolumn{1}{l|}{}  & \multicolumn{1}{l|}{}  & \multicolumn{1}{l|}{}  &    &  14.7$\pm$0.3511                   \\ \cline{2-8}
                         & \multicolumn{1}{l|}{}  & \multicolumn{1}{l|}{\checkmark}  & \multicolumn{1}{l|}{}  & \multicolumn{1}{l|}{}  & \multicolumn{1}{l|}{}  &    &  14.2$\pm$0.3417                  \\ \cline{2-8}
                         & \multicolumn{1}{l|}{}  & \multicolumn{1}{l|}{}  & \multicolumn{1}{l|}{\checkmark}  & \multicolumn{1}{l|}{}  & \multicolumn{1}{l|}{}  &    &  13.9$\pm$0.2283                  \\ \cline{2-8}
                         & \multicolumn{1}{l|}{}  & \multicolumn{1}{l|}{}  & \multicolumn{1}{l|}{}  & \multicolumn{1}{l|}{\checkmark}  & \multicolumn{1}{l|}{}  &    &  14.5$\pm$0.4314                   \\ \cline{2-8}
                         & \multicolumn{1}{l|}{}  & \multicolumn{1}{l|}{}  & \multicolumn{1}{l|}{}  & \multicolumn{1}{l|}{}  & \multicolumn{1}{l|}{\checkmark}  &    &  14.3$\pm$0.6315                   \\ \cline{2-8}
                         & \multicolumn{1}{l|}{}  & \multicolumn{1}{l|}{}  & \multicolumn{1}{l|}{}  & \multicolumn{1}{l|}{}  & \multicolumn{1}{l|}{}  & \checkmark   &  14.6$\pm$0.1354         \\ \cline{2-8}
                         & \multicolumn{1}{l|}{\checkmark}  & \multicolumn{1}{l|}{}  & \multicolumn{1}{l|}{}  & \multicolumn{1}{l|}{\checkmark}  & \multicolumn{1}{l|}{}  & \checkmark   &            \textbf{15.4}$\pm$\textbf{0.3215}         \\ \hline
\end{tabular}
\end{table}

\begin{figure}
\centering
\includegraphics[width=0.8\linewidth]{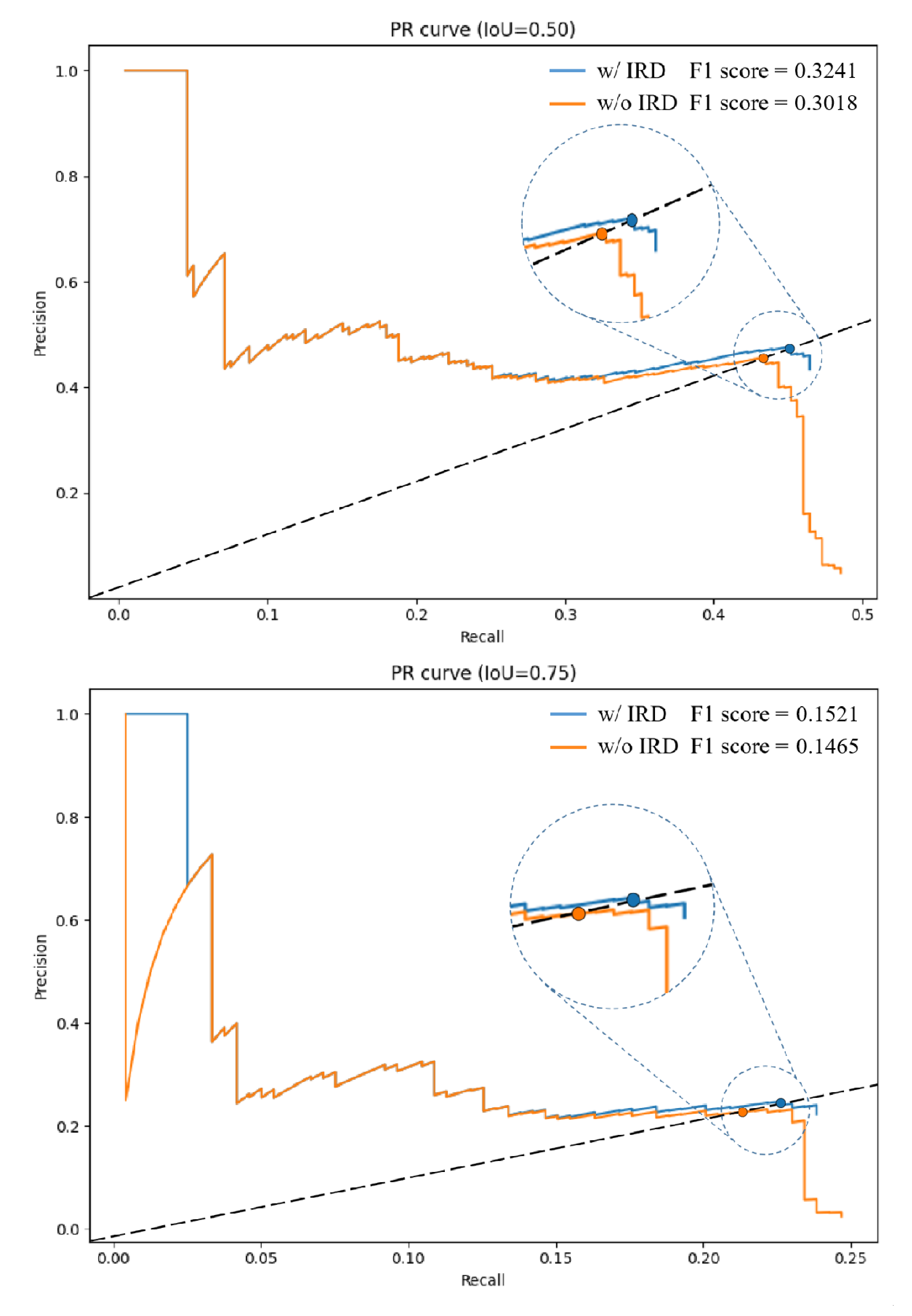}
\caption{PR curve of the TSIPR in different IoU.} \label{fig_pr}
\end{figure}

\subsubsection{Component Ablation}
To validate the effectiveness of the proposed multi-scale patch embedding module, we compare the network performance of both single-scale and multi-scale patch embedding. Specifically, we test the patch size from 5 to 10 in the experiment and the results are listed in Table~\ref{table_patchsize}. From the table we can observe that the performance of the single-scale network with different patch sizes is similar. We choose three patch sizes (5, 8, 10) with the highest performance for our multi-scale patch embedding module. As the combination of multi-scale patch embedding can extract robust features from images with different tooth spacing, the multi-scale patch embedding can improve nearly $0.7\thicksim1.5\%$ location performance compared to the single scale one. This experimental results are consistent with our assumption and demonstrate the effectiveness of the proposed multi-path patch embedding method.

In Fig.~\ref{fig_exp_patch}, we visualize the detection results of the single-scale and multi-scale patch embedding network. The visualization indicates that the multi-scale patch embedding network generate more accurate detection results than the single-scale.

\subsubsection{Ablation of the GLFIB}
To demonstrate the effectiveness of the proposed GLFIB, we conduct an ablation of convolution and transformer in the GLFIB. Results are listed in Table~\ref{table_GLFIB}. From the table we can observe that the pure transformer architecture achieves an AP of 14.0\%, which outperforms the pure convolutional architecture by 1.8\% AP. When introducing a convolution into the GLFIB, the AP value improves by 1.4\%. This experimental result demonstrates that the extracted local texture from the CNN can provide fine-grained feature for the prediction network. With the increment of the number of convolutions, the AP value decreases. This phenomenon illustrates that the prediction of implant position relies on the global context, which is consistent with our design intention. The experimental results validate the effectiveness of the GLFIB, which combines the convolution and transformer for better feature extraction.

\begin{figure}
\centering
\includegraphics[width=0.8\linewidth]{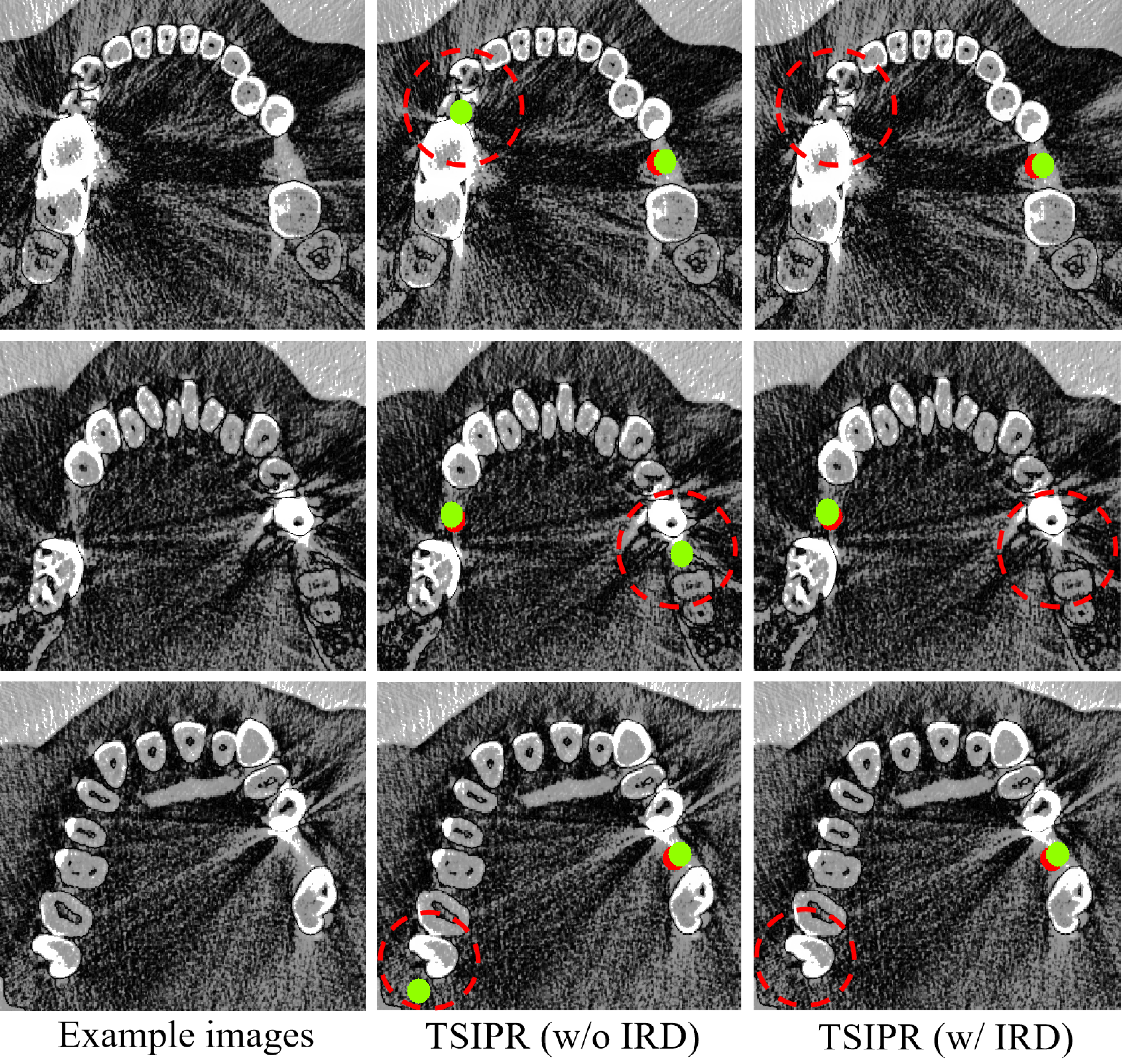}
\caption{Comparison of the detection results of the MSPENet and TSIPR. The red and green circles denote the ground-truth position and the predicted implant position, respectively. The dashed red circle indicates the region of false detection.} \label{fig_ablation}
\end{figure}

\begin{table}
\centering
\caption{Ablation study of the proposed GLFIB. T and C denote the transformer and convolution, respectively.}\label{table_GLFIB}
\begin{tabular}{|l|l|l|}
\hline
Network        & GLFIB        & $AP_{75}\%$  \\ \hline
\multirow{5}{*}{MSPENet} & {[}T,T,T,T{]} & 14.0$\pm$0.4713     \\ \cline{2-3}
                         & {[}C,T,T,T{]} & \textbf{15.4}$\pm$\textbf{0.3215}      \\ \cline{2-3}
                         & {[}C,C,T,T{]} & 14.6$\pm$0.6135     \\ \cline{2-3}
                         & {[}C,C,C,T{]} & 13.3$\pm$0.3354     \\ \cline{2-3}
                         & {[}C,C,C,C{]} & 12.2$\pm$0.5241     \\ \hline
\end{tabular}
\end{table}

\subsubsection{Branch Ablation}
As previously discussed, MSPENet might easily generate false detection at the sparse teeth region. We conducted an ablation experiment to validate whether the proposed IRD can reduce the false detection rate. Fig.~\ref{fig_pr} shows the PR curve and the F1 score of the TSIPR for different IoU, the abscissa and ordinate of PR curve is Recall and precision, respectively; on the dash line the recall equals to precision. From the curves we can observe that for both IoU of 0.5 and 0.75, the PR curve of network with IRD branch is above the baseline PR curve, which indicates that the IRD branch can improve the detection performance on both recall and precision. The F1 score of the network with IRD are also 2.23\% and 0.56\% higher when IoU equals to 0.5 and 0.75, respectively.

We also visualize the detection results of the MSPENet and TSIPR in Fig.~\ref{fig_ablation}. We can observe from the figure that the detection results predicted by the MSPENet have false detections in the teeth area with large space. When the IRD is introduced, the false detection results are filtered greatly, which is consistent with the experimental results.

\begin{table}
\caption{Comparison of the proposed method with other mainstream detectors.}\label{table_detector}
\centering
\scalebox{0.77}{
\begin{tabular}{|l|l|l|l|l|l|l|}
\hline
Methods                            & Network         & Backbone             & $AP_{75}\%$                    & F1 Score                       & Param(M)        & FPS \\ \hline
\multirow{5}{*}{CNN-based}         & CenterNet       & \multirow{6}{*}{R50} & 10.9$\pm$0.2                   & 10.8$\pm$0.1                   & 32.29           & \textbf{69}  \\ \cline{2-2} \cline{4-7}
                                   & ATSS            &                      & 12.1$\pm$0.2                   & 11.9$\pm$0.2                   & 32.30           & 35  \\ \cline{2-2} \cline{4-7}
                                   & VFNet           &                      & 11.8$\pm$0.8                   & 11.8$\pm$0.1                   & 32.78           & 28  \\ \cline{2-2} \cline{4-7}
                                   & RepPoints       &                      & 11.2$\pm$0.1                   & 11.1$\pm$0.3                   & 36.84           & 16  \\ \cline{2-2} \cline{4-7}
                                   & ImplantFormer   &                      & 11.5$\pm$0.3                   & 11.3$\pm$0.3                   & 24.73           & 65  \\ \cline{1-2} \cline{4-7}
\multirow{3}{*}{Transformer-based} & Deformable DETR &                      & 12.8$\pm$0.1                   & 12.5$\pm$0.1                   & 41.07           & 22  \\ \cline{2-7}
                                   & ImplantFormer   & ViT-B-R50            & 13.7$\pm$0.2                   & 13.6$\pm$0.2                   & 100.52          & 14  \\ \cline{2-7}
                                   & MSPENet(ours)   & \multirow{2}{*}{-}   & 15.4$\pm$0.3                   & 15.2$\pm$0.2                   & \textbf{14.21}  & 58  \\ \cline{1-2} \cline{4-7}
-                                  & TSIPR(ours)     &                      & \textbf{15.7}$\pm$\textbf{0.4} & \textbf{15.6}$\pm$\textbf{0.3} & 85.51           & 46  \\ \hline
\end{tabular}
}
\end{table}

\subsubsection{Comparison to the mainstream Detectors}
To demonstrate the superiority of our method, we compare the location performance of the proposed TSIPR with other mainstream detectors. As little useful texture is available around the center of implant, the anchor-based detectors cannot regress implant position successfully. Only the CNN-based anchor-free detectors (VFNet~\citep{zhang2021varifocalnet}, ATSS~\citep{zhang2020bridging}, RepPoints~\citep{yang2019reppoints}, CenterNet~\citep{zhou2019objects}), transformer-based detectors (Deformable DETR~\citep{zhu2020deformable} and ImplantFormer~\citep{yang2022implantformer}) are employed for comparison. Results are listed in Table~\ref{table_detector}.

From the table we can observe that the transformer-based methods perform better than the CNN-based networks (e.g., ImplantFormer achieved 13.7\% AP, which is 1.6\% higher than the best performed CNN-based network - ATSS). This experimental result demonstrates that the capacity of building the long-ranged dependency is important for the implant position regression. Our method, MSPENet, achieves the highest AP - 15.4\% among the transformer-based methods, which outperforms the ImplantFormer by 1.7\% AP. When applying the IRD to filter out the false detection, the AP value reaches 15.7\%.

To further validate the effectiveness of TSIPR, we introduce more metrics for comparison, i.e., F1 score, parameter, and FPS. From the table we can observe that the proposed TSIPR also performs the best, in terms of F1 score, with reasonable efficiency. These experimental results prove the effectiveness of our method, which achieves the best performance among all benchmarks.

\section{Conclusion and Future Work}\label{sec13}
In this study, we develop a two-stream implant position regression framework (TSIPR) for CBCT data based implant position prediction, which consists of an implant region detector (IRD) and a multi-scale patch embedding regression network (MSPENet). We extend the original annotation to provide additional supervisory information for the training of IRD, which locates the most probable bounding box of implant region to filter out the false regressions generated by the MSPENet. For the MSPENet, a multi-scale patch embedding module is designed to adaptively extract features from the images with various tooth spacing. The global-local feature interaction block is designed to build the encoder of MSPENet, which combines the transformer and convolution for enriched feature representation. Extensive experiments on a dental implant dataset demonstrated that the proposed TSIPR achieves superior performance than the existing methods, especially for patients with sparse teeth.

Although the proposed TSIPR achieves a promising performance than the previous methods, it has some limitations. Firstly, the annotation of the implant position is difficult, which requires a pair of CBCT data captured pre- and post-implantation, for each patient. Secondly, TSIPR does not fully explore 3D context. As the network input is a single 2D slice of tooth crown, the texture variation between the neighbored slices is not used. Therefore, in the future work, we will explore semi-supervision approaches and expand the TSIPR to take multiple slices as input to fully explore 3D context. In real clinics, as IRD only output the most probable implant region, TSIPR can not perform well for patients with multiple missing teeth.

\section*{Acknowledgments}
This work was supported by the National Natural Science Foundation of China under Grant 82261138629; Guangdong Basic and Applied Basic Research Foundation under Grant 2023A1515010688 and 2021A1515220072; Shenzhen Municipal Science and Technology Innovation Council under Grant JCYJ20220531101412030 and JCYJ20220530155811025.



\bibliography{ref}

\end{document}